\documentclass[10pt,twocolumn,letterpaper]{article}

\usepackage[pagenumbers]{wacv} 

\usepackage{multirow}
\usepackage{times}
\usepackage{epsfig}
\usepackage{epigraph}
\usepackage{enumitem}
\usepackage{float}
\usepackage{chngcntr}
\usepackage{soul}
\usepackage{caption}
\usepackage{subcaption}
\usepackage{setspace} 
\usepackage{zref-xr}
\usepackage{colortbl}
\usepackage{pifont}

\usepackage{graphicx}
\usepackage{amsmath}
\usepackage{amssymb}
\usepackage{booktabs}

\usepackage[pagebackref,breaklinks,colorlinks]{hyperref}

\usepackage[capitalize]{cleveref}
\crefname{section}{Sec.}{Secs.}
\Crefname{section}{Section}{Sections}
\Crefname{table}{Table}{Tables}
\crefname{table}{Tab.}{Tabs.}
\definecolor{baselinecolor}{gray}{.8}
\definecolor{teal}{rgb}{0.0, 0.5, 0.5}

\def\wacvPaperID{***} 
\def\confName{WACV}
\def\confYear{2024}

\begin{document}

\title{Learning Visual Grounding from Generative Vision and Language Model} 

\author{Shijie Wang$^{1*}$ \and Dahun Kim$^{2}$ \and Ali Taalimi$^{2}$ \and Chen Sun$^{1}$ \and Weicheng Kuo$^{2}$}

\author{
Shijie Wang$^{1}$\thanks{Work done while at Google DeepMind.}
\qquad
Dahun Kim$^{2}$
\qquad
Ali Taalimi$^{3}$
\qquad
Chen Sun$^{1}$
\qquad
Weicheng Kuo$^{2}$
 \\
$^{1}$ Brown University \quad $^{2}$ Google DeepMind \quad $^{3}$ Google Cloud
}

\maketitle
\begin{abstract}
Visual grounding tasks aim to localize image regions based on natural language references. In this work, we explore whether generative VLMs predominantly trained on image-text data could be leveraged to scale up the text annotation of visual grounding data. We find that grounding knowledge already exists in generative VLM and can be elicited by proper prompting. We thus prompt a VLM to generate object-level descriptions by feeding it object regions from existing object detection datasets. We further propose attribute modeling to explicitly capture the important object attributes, and spatial relation modeling to capture inter-object relationship, both of which are common linguistic pattern in referring expression. Our constructed dataset (500K images, 1M objects, 16M referring expressions) is one of the largest grounding datasets to date, and the first grounding dataset with purely model-generated queries and human-annotated objects. To verify the quality of this data, we conduct zero-shot transfer experiments to the popular RefCOCO benchmarks for both referring expression comprehension (REC) and segmentation (RES) tasks. On both tasks, our model significantly outperform the state-of-the-art approaches without using human annotated visual grounding data. Our results demonstrate the promise of generative VLM to scale up visual grounding in the real world. Code and models will be released.

\end{abstract}
    
\section{Introduction}
Visual grounding aims to identify image region described by a natural language query. It provides a strong foundation for tasks like visual reasoning and human-robot interaction. One of the most popular visual grounding task referring expression comprehension (REC)~\cite{refcoco, refcocog} aims to localize an object by language references and generate the corresponding bounding box. Another similar task is referring expression segmentation (RES), which requires pixel-level grounding to generate the referred object mask. Visual grounding tasks require models to accurately understand the natural language descriptions of the referred objects' categories, visual appearance, and their relationships with other objects or the scene. An important challenge is the limited scale of grounding datasets: Traditional visual grounding datasets like~\cite{kazemzadeh2014referitgame, refcoco,refcocog,nagaraja2016modeling} require detailed human labeling and verification to generate object-level text annotations, making it expensive and unscalable in practice. For example, the popular RefCOCO/+/g REC datasets only contain $\mathcal{O}$(10K) images and objects. In contrast, detection datasets~\cite{lin2014microsoft,gupta2019lvis,shao2019objects365,kuznetsova2020open} are much larger and more diverse with $\mathcal{O}$(1M) images and $\mathcal{O}$(1-10M) objects. Such 2-3 orders of magnitude difference show the promise of learning grounding from large-scale detection datasets.

Generative Vision-Language Models (VLMs)~\cite{alayrac2022flamingo, chen2023pali, chen2023pali3, li2023blip} flexibly unify multiple vision-language tasks such as visual-question answering (VQA) and image captioning without task-specific design by formatting various tasks as image-and-text to text generation. Most current VLMs~\cite{alayrac2022flamingo,li2023blip,chen2023pali,lu2023unifiedio} are primarily pre-trained on large image-text datasets~\cite{zhai2022scaling, schuhmann2022laion, chen2023pali} for natural language generation tasks. They lack the ability to directly perform localization tasks (\ie generating boxes or masks) without downstream fine-tuning. This raises the question of whether these VLMs inherently lack object-level knowledge and grounding capability.
Conceptually, both scene-level and object-level visual information can be encapsulated within an image at various scales (\eg an image of a forest and a tree). Practically, generative VLMs are usually trained on $\mathcal{O}$(1B) datasets that contain feature diverse distribution covering complex scenes and object-centric images. We thus hypothesize that VLMs can learn object-centric knowledge solely from image-level pre-training tasks and produce object-level descriptions when ``zooming-in” an image to the single object region, which can be used for grounding model training. Consequently, without expensive fine-tuning on grounding dataset, a generalist generative VLM may naturally serve as a teacher model, providing supervision for student visual grounding models without extra human annotations.

As motivated above, we utilize PaLI-3~\cite{chen2023pali3}, a state-of-the-arts generative VLM to construct large-scale referring expression dataset automatically for scalable visual grounding. To better model human linguistic patterns beyond single-object descriptions, we also employ spatial relationship heuristics and attributes priors to enrich the generated region-text dataset. We create VLM-VG, a large-scale dataset for fine-grained visual grounding created based on two object detection datasets: COCO 2017~\cite{lin2014microsoft} and Objects365 v1~\cite{shao2019objects365}, which are 1-2 orders of magnitude larger than existing grounding datasets. We focus on the ``zero-shot” referring expression comprehension/segmentation setting, where no human-annotated text annotations are utilized during pre-training. We pre-train a simple light-weight grounding model on the proposed dataset and perform zero-shot evaluation on the RefCOCO/+/g datasets, conducting systematic comparisons and analyses. Benefiting from the large-scale grounding pre-training with high-quality referring expression annotations from generative VLMs and relation modeling, we significantly outperform prior zero-shot REC/RES methods. In summary, our work has three main contributions:

\begin{itemize}
\item[$\bullet$] We hypothesize and empirically observe that image-text pre-trained generative VLMs can naturally provide high-quality object-level captions for single-object regions. In this way, we leverage a generalist VLM to automatically generate referring expression annotations to provide supervision for specialist visual grounding models, thereby overcoming the limitations of small-scale grounding datasets.

\item[$\bullet$] We further enrich the auto-labeled region-text pairs by leveraging heuristics of spatial relationships and cross-domain knowledge of object attributes, providing more comprehensive and diverse query annotations.

\item[$\bullet$] We introduce VLM-VG, a visual grounding dataset for scalable referring expression comprehension/segmentation, without requiring manual grounding annotations. By pre-training on VLM-VG, we achieve SoTA zero-shot performance on RefCOCO/+/g benchmarks in both REC and RES tasks with a lightweight Faster R-CNN based model. 

\end{itemize}

\section{Related Work}
 \begin{figure*}[tb]
    \centering
    \includegraphics[width=\textwidth]{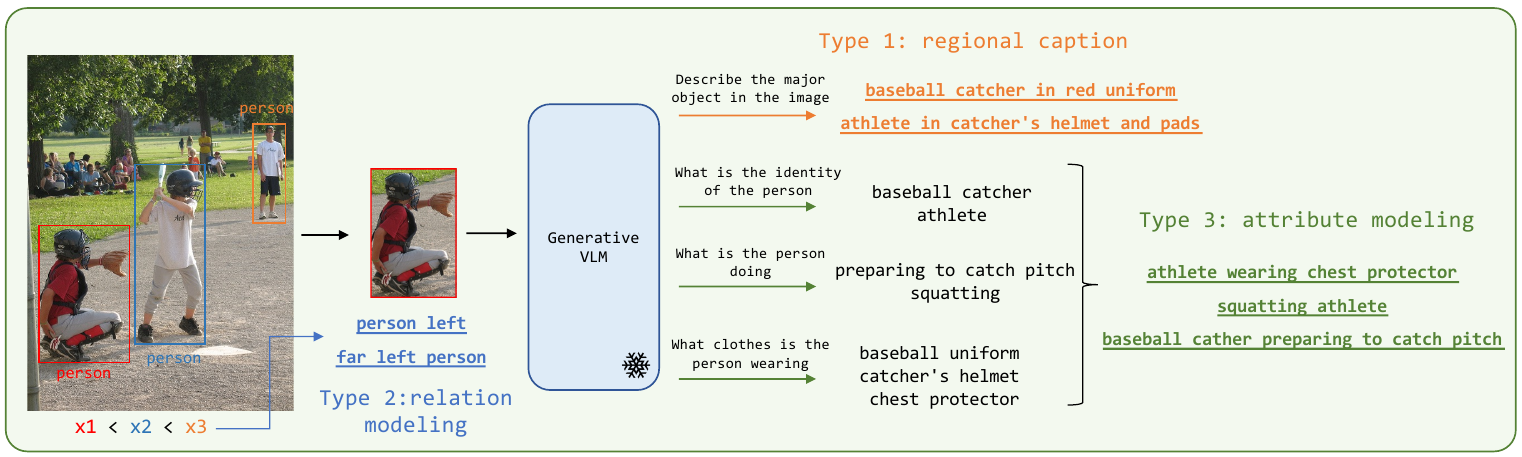}
    \captionsetup{width=\textwidth} 
    \caption{
    \textbf{Overview of referring expression generation.} We propose to generate grounding annotations automatically using generative vision language models. We construct three types of annotations to model the diversity of human linguistic norms. Type 1: regional caption directly generated by prompting the VLM with a generic instruction to describe the major object in the cropped image regions. Type 2: relational descriptions generated by rule-based methods utilizing spatial heuristics. Type 3: Attribute-rich descriptions that explicitly model attributes of the object by querying the VLM with ``attribute” prompts.
    }
    \label{fig:model}
\end{figure*}

\noindent \textbf{Visual Grounding} is a critical ability for AI systems. Specifically, it aims to identify and associate specific parts of visual data with corresponding textual descriptions, thus bridging the semantic gap between the high-dimensional visual representations and the symbolic language representations. Referring expression comprehension/segmentation~\cite{refcoco,refcocog, liu2019clevr, liu2023gres} (REC/RES), and phrase grounding~\cite{plummer2015flickr30k, kazemzadeh2014referitgame, vg, wu2020phrasecut} (PG) are two common visual grounding tasks. REC/RES tasks require the model to detect or segment a single object from the image based on language queries referring to the object while PG aims to localize each entity mentioned by the phrase. In this work, we focus on the referring expression comprehension and segmentation task requiring fine-grained object-level understanding capabilities. Compared with other localization tasks such as detection dataset~\cite{lin2014microsoft,shao2019objects365,kuznetsova2020open}, grounding datasets~\cite{refcoco,refcocog} are much smaller due to the high annotation cost.

\noindent \textbf{Generative Vision Language Models} has made tremendous advances across various areas. Inspired by token masking~\cite{devlin-etal-2019-bert} or next-token prediction~\cite{brown2020language} training objectives utilized by text modeling, generative vision-language pre-training~\cite{wang2022git,chen2023pali,wang2021simvlm,wang2022image,chen2020uniter,UnifiedVLP,VinVL,vilbert2020,Su2020VL-BERT,lu2023unifiedio, huang2024language} extends similar ideas to various tasks that can be modeled by a general ``image-text to text” interface. Optimized by a single language modeling loss, large-scaled generative vision-language models can be flexibly applied to multiple image-language downstream tasks including image captioning, visual question answering (VQA), or image classification. Recent works~\cite{chen2023shikra, kosmos2, you2024ferret, li2024covlm, xu2024pixel} show VLMs can ground objects after large-scale region-text fine-tuning, which could be expensive and resource-consuming. Motivated by this, we investigate whether image-level generalist VLMs can ground objects without heavy downstream fine-tuning. We observe that VLMs could provide informative object-level captions given a single-object image regions. The generated regional captions can provide robust and generalizable supervision for specialist grounding models, enabling us to scale up visual grounding datasets for free.

\noindent \textbf{Unsupervised Visual Grounding} can be categorized into two types. One type of works~\cite{yao2021cpt, subramanian2022reclip, han2023zero, shtedritski2023does, yu2023zero, liu2023vgdiffzero, ni2023ref} leverage pre-trained visual-language models (VLMs) such as CLIP~\cite{clip}, FLAVA~\cite{singh2022flava}, or Stable Diffusion~\cite{stablediffusion} with strong visual language alignment capability to retrieve box or mask from pre-extracted box/mask proposals generated by off-the-shelf detectors~\cite{ren2015faster} or segmentation models~\cite{he2017mask,kirillov2023segment}. Another line of works~\cite{li2021grounded, liu2023grounding, zhao2024open, kosmos2} take in images directly and pre-train the grounding model on detection~\cite{lin2014microsoft,shao2019objects365} and grounding~\cite{vg,mdetr} datasets, then perform zero-shot evaluation on the down-stream tasks. Specifically, ~\cite{li2021grounded, liu2023grounding, zhao2024open} combine base detection models with visual-language fusion modules while~\cite{kosmos2, li2024covlm} utilize large language models to formulate REC task as language modeling.

We adopt the second way to pre-train a simple grounding model following~\cite{kuo2022findit} on our constructed grounding dataset, VLM-VG. Different from~\cite{jiang2022pseudo} which constructs pseudo-queries by parsing images with object and attribute detectors and composing attributes into on phrases by rule-based templates, and~\cite{kosmos2}, which isolates object-related nouns from image-level captions to construct referring expressions, we propose to leverage generative visual-language models to construct language annotations automatically for object-centric regions in the images by using box annotations from detection dataset. In this way, we leverage detection datasets that are generally 1-2 orders of magnitude larger in terms of images and objects and bypass the requirements of human annotations.
\section{Method}
\subsection{Generative VLM}

Generative vision language models take image and text as input and generate text by autoregressive token prediction or masked token prediction. Since a variety of tasks can be converted into text generation tasks by designing different instructions and prompts, generative VLMs can thus handle multiple tasks without task-specific training objectives. Formally, a generative VLM takes in the image-text pair $\texttt{<image, prompt>}$, where different downstream tasks can be solved by different $\texttt{prompt}$ designs. For instance, in the Visual Question Answering (VQA) task, the prompt can be constructed as a combination of the question and answering instructions. By prompting the generative generalist VLMs, it offers a unified interface to solve tasks directly and provide knowledge and information for other models.

\subsection{Referring Expression Generation}
\label{sec:dataset_construct}
In referring expression comprehension/segmentation tasks, the language queries generally focus on two key factors: accurate descriptions of objects' visual appearance and relationship with other objects in the images. Taking both into account, we propose to automatically generate large-scale referring expressions combining three types of queries. As demonstrated in Figure~\ref{fig:model}, we describe the referring expression generation process below:

\noindent\textbf{Type 1: Regional Caption.} Given an image containing multiple instances/objects represented by corresponding ground-truth bounding boxes, the goal is to generate descriptive language queries referring to the objects. For each image, we first filter out small objects whose area accounted for less than a threshold ratio $K$ (\ie $K$=0.05) of the entire image area. For the remaining objects, we crop the image using the associated bounding boxes to generate single object regions to isolate the objects from the influence of the context. Then we leverage a VLM to generate captions providing informative descriptions for the center object. In detail, we feed the object-centric regions to a state-of-the-arts generative VLM, PaLI-3~\cite{chen2023pali3}, and prompt it with language instructions following the image caption task. Since the cropped image may still include irrelevant object parts, we empirically design the prompt to instruct the model to focus on the center object and ignore the irrelevant background: "\texttt{Describe the major object in the image, ignore the background.}". Taking in the image-text pair of the cropped regions and language instructions, the VLM generates multiple predictions with corresponding confidence. We choose the top-5 answers with highest confidence as the expressions referring to the corresponding object without any further post-processing.

\noindent \textbf{Type 2: Relation Modeling.}
The type 1 regional captions provide ``local“ descriptive references by isolating the center object from the whole image. However, the single object regions filter out global information such as positions in the scene and relationship with other objects, which is also a crucial perspective to refer to an object/instance. For example, in an image containing two cups with similar appearance positioned in different locations, it hard to successfully locate one cup solely relying on the regional descriptions on its visual appearance. We observe that spatial relationship is one of the most common relations in human cognition process to distinguish objects. Inspired by previous works~\cite{subramanian2022reclip, jiang2022pseudo}, we design a rule-based method to model spatial relationship between objects utilizing bounding boxes as localization heuristics on three dimensions: horizontal, vertical and depth.

We model both relative and absolute spatial relationships which can be formulated by the tuple \texttt{(noun, rel, noun)} and \texttt{(noun, rel)} respectively. We use the object category label of detection dataset as the \texttt{noun} \eg ``person”. To model the spatial relation, we treat the center coordinate of the bounding box as the position of an object. On the horizontal dimension, we compare one object with other objects by the x-coordinates for relative position. For absolute position, we normalize an object's x-coordinate against the image width, categorizing locations as ``left” (\texttt{[0, 0.25)}), ``center/middle” (\texttt{[0.25, 0.75]}), and ``right” (\texttt{(0.75, 1]}). Vertically, we solely focus on absolute positioning through y-coordinate comparison. Similarly, we normalize the y-coordinates and categorize horizontal location as ``top”(\texttt{[0, 0.25)}) and ``bottom” (\texttt{(0.75, 1]}). For depth, following~\cite{jiang2022pseudo} that assumes objects with larger area is more likely to be closer to the camera (\ie ``front”) and we only model the absolute position. We first calculate the ratio of the area of the smallest object to the largest object, we model depth relation for this image only if the ratio is smaller than a threshold (\ie 0.4), which denotes there's significant scale differences among the objects. Upon satisfying this condition, we categorize an object's depth position by the the ratio of its area to the largest object: ``behind” (\texttt{[0, 0.4)}) and ``front” (\texttt{(0.8, 1]}). 
Following the rule-based methods, the extracted relation tuples are then formulated into referring phrases: "\texttt{noun rel noun}" for relative relation, "\texttt{rel noun}" or "\texttt{noun rel}" for absolute relation to expand the diversity of referring expressions.

The rule-based spatial relation modeling provides spatial-aware annotations absorbing information from localization heuristics. But it still has some limitations. For example, to determine objects' location solely based on its center point may fail especially when two objects are close or have occlusion, and using box area to categorize the depth is just a rough approximation since different objects might have various intrinsic sizes.

\noindent \textbf{Type 3: Explicit Attribute Modeling.}
Attributes are important ways humans refer to objects. However, the generic region captions generated by prompting the VLM with a ``general prompt” from type 1 may not contain adequate attribute information or may dilute the attribute information with other co-occurring words. We further ask the question: Will explicitly modeling object's attributes further complement the potential missing information in the generic captions and match human referring expression better?

We then propose to leverage both Large Language Model and Vision and Language Model to construct attribute-rich descriptions. Specifically, we first prompt GPT-4~\cite{achiam2023gpt} to enumerate common attributes to identify an object with the prompt "\texttt{what are the common attributes to identify an object?}" and then manually pick 7 typical attributes: \textit{[cloth, action, gender, identity, color, material, shape]}. For each object category in the dataset, we further utilize GPT-4 to pick applicable attributes from the 7 candidate attributes. In this way, we associate each object class with its related attributes. We design a prompt template for each attribute (\eg for \textit{color}, the prompt is formulated as "\texttt{What is the color of the [object class]}"). The detailed prompt templates are listed in Sec.~\ref{supp:details}. We then query PaLI-3 with the cropped region and attribute prompt to get the attribute of the center object by choosing the top-3 answers. Interestingly, we observe sometimes PaLI-3 will generate ``unkown” or ``unsuitable“ when the attributes is not applicable to the object class. In this case, we just ignore this attribute for the specific object. Finally, after tagging each object with attributes, the attribute-rich descriptions are then formulated by combining attribute tags in the format of "\texttt{noun adj} or "\texttt{adj noun}" randomly, where we define \textit{gender, identity} and the class label as the nouns, and \textit{cloth, action, color, material, shape} as the adjectives.

\begin{table*}[t]
\centering
\resizebox{0.8\textwidth}{!}{
\begin{tabular}{lcccccc}
\toprule
\bf Dataset & Images & Objects & Text &Data Source & Box Anno. & Text Anno. \\ \midrule
ReferIt~\cite{referitgame} &20K &97K &131K &ImageCLEF &Human &Human\\
RefCOCO~\cite{refcoco} & 32K & 50K & 142K &COCO &Human &Human \\
RefCOCO+~\cite{refcoco} & 20K & 50K & 142K &COCO &Human &Human \\
RefCOCOg~\cite{refcocog} & 27K & 55K & 85K &COCO &Human &Human \\ 
Flickr Entities~\cite{plummer2015flickr30k} & 31K & 275K & 513K & Flickr30k &Human &Human \\
Visual Genome~\cite{vg} & 108K & 4.1M & 4.1M &COCO & Human & Human \\
\textsc{GrIT}~\cite{kosmos2} &  91M & 137M & 115M &COYO-700M, LAION-2B & Model & Web + Model \\
\textsc{COVLM-97M}~\cite{kosmos2} &  97M & - & - &BLIP-2 & Model & Web + Model \\
\midrule
\bf VLM-VG-COCO  &81K  &143K  &2M  & COCO$^*$ &Human & Model\\
\bf VLM-VG-O365  &431K   &958K  &14.2M  & O365 &Human & Model\\
\bf VLM-VG (Ours) &512K   &1.1M  &16.2M  & COCO$^*$, O365 &Human & Model\\
\bottomrule
\end{tabular}
}
\vspace{-3mm}	
\caption{Compare VLM-VG with existing visual grounding datasets. COCO$^*$ denotes the subset of COCO that excludes images used by RefCOCO/+/g. Different from all the existing grounding datasets, our VLM-VG datasets uses pure model model-genereated text annotations without relying on manual text labeling.
}
\label{tab:dataset}
\end{table*}

\subsection{VLM-VG: From Detection to Grounding}
Following the steps above and combing the three types of annotations, we construct our dataset, VLM-VG, based on the training set of two object detection datasets COCO 2017~\cite{lin2014microsoft} and Objects365 v1~\cite{shao2019objects365}. Specifically, for COCO, we filter out images present in the RefCOCO/+/g validation and test splits to avoid contamination during evaluation. For the Objects365 dataset, we keep the object categories corresponding to the 80 COCO classes to reduce the computation cost. For the vision-language model, we leverage the 5B-parameter PaLI-3 model which was fine-tuned on image captioning and visual question answering tasks, as detailed in~\cite{chen2023pali3}.

In Table~\ref{tab:dataset}, we compare VLM-VG with existing visual grounding datasets in terms of size, data sources, and annotation methods. Traditionally, most visual grounding datasets rely on object-level text queries labeled by human annotators~\cite{referitgame,refcoco,refcocog,plummer2015flickr30k,vg}, which is associated with high labor costs, limited flexibility and scalability. Two recent datasets \cite{kosmos2, li2024covlm} adopt image-caption pairs from web-collected datasets\cite{schuhmann2022laion, kakaobrain2022coyo-700m} and utilizes GLIP~\cite{li2021grounded} trained on Objects365~\cite{shao2019objects365} and GoldG~\cite{mdetr} to generate object boxes and uses NLP tools~\cite{honnibal2020spacy} to isolate object-level word chunks from image-level captions and expand them into referring expressions using dependency relations of the caption sentences. Despite its scale, this method is constrained by the quality of the web image captions (\eg alt-text) and struggles to provide detailed and diverse references for objects from various perspectives that align with human linguistic norms.

In contrast, VLM-VG extends existing detection datasets for visual grounding by constructing referring expressions using generative vision language models, without the need of human annotations. Notably, VLM-VG is the first dataset where the texts are purely annotated by model with high quality region-text alignment. Sourcing from COCO and Objects365, VLM-VG contains 512K images, 1.1M objects and 16.2M referring expressions in total, averaging 14.7 referring expressions per object. This abundance of comprehensive and diverse descriptions facilitates a more accurate modeling of real-world referring expression, thereby enhancing the effectiveness and generalizability of visual grounding models in complex practical scenarios.

\subsection{Visual Grounding: Task and Model}
To evaluate the quality of the generated dataset, we utilize a simple and lightweight model proposed by~\cite{kuo2022findit} on two visual grounding tasks: referring expression comprehension (REC) and segmentation (RES). In the REC/RES task, the model takes in an image and a language query about a specific object in the image to generate \textbf{one} bounding box/mask of the referred object.

The model adopts a simple and extensible architecture design: it includes a ResNet~\cite{he2016deep} as the image encoder to produce multi-level features, T5~\cite{raffel2020exploring} as the text encoder, a multi-level multimodal fusion network~\cite{kuo2022findit}, and box/mask prediction heads~\cite{he2017mask}. The fusion network uses cross-attention and feature product to fuse the multi-level image and text features. After the vision-language fusion, a standard region proposer~\cite{ren2015faster} and a box/mask prediction heads are applied to decode the predicted box/mask~\cite{kuo2022findit}. During training, the model is optimized by the box regression loss and optionally mask loss~\cite{he2017mask}. Following the common practice, we initialize the ResNet~\cite{he2016deep} backbone with checkpoints pretrained on the COCO detection task and the T5 encoder with language modeling pretrained checkpoint. We use a batch size of 1024, image size 384$\times$384, total training steps 200K, Momentum optimizer with initial learning rate 0.04 and step decay at 70\% and 90\% of total steps by a factor of 0.1 and linear warmup of 500 steps. Due to the initialization, we set both image and text encoder learning rate to 0.1 of the initial learning rate. We also apply image scale jittering with a factor $S$ uniformly sampled from $S \sim Uniform(0.5, 2.0)$ at training time.

For the RES task, since the segmentation datasets are generally less scalable than detection datasets, we propose to treat REC as the pre-training task of RES and initialize the grounding with our pre-trained REC models. During RES training, only mask head is fine-tuned while the rest of the model is frozen. We use batch size 64, learning rate 0.01 and 50K training steps for the RES task.

\section{Experiments}

\begin{table*}[t]\footnotesize
\begin{center}
\resizebox{2\columnwidth}{!}{
\begin{tabular}{l|c|c|ccc|ccc|cc|c}
    \toprule
    \multirow{2}{*}{Method} &\multirow{2}{*}{Vison Backbone} & \multirow{2}{*}{Pretraining Dataset} &  \multicolumn{3}{c|}{RefCOCO} & \multicolumn{3}{c|}{RefCOCO+} & \multicolumn{2}{c|}{RefCOCOg} & \multicolumn{1}{c}{Avg}\\
    & & & val & testA & testB & val & testA & testB & val & test &\\

    \midrule
CPT~\cite{yao2021cpt}  &ResNeXt-152 &\textcolor{blue}{OpenImages}, \textcolor{blue}{COCO}, \textcolor{blue}{O365}, \textcolor{blue}{VG} & 32.2 & 36.1 & 30.3 & 31.9 & 35.2 & 28.8 &36.7 &36.5  &33.5\\
ReCLIP$^*$~\cite{subramanian2022reclip} &ViT-B+R50 &CLIP, \textcolor{blue}{COCO} & 45.8 & 46.7 & 45.2 & 45.3 & 48.5 & 42.7 &57.0 &56.2  &48.4\\
Red Circle$^*$~\cite{shtedritski2023does} &ViT-L+R50 & CLIP,\textcolor{blue}{COCO} & 49.8 & 58.6 & 39.9 & \bf55.3 & \bf63.9 & 45.4 & 59.4 & 58.9 & 53.9\\
RelVLA~\cite{han2023zero}  &ViT-B & PMD, \textcolor{blue}{HICO}, \textcolor{blue}{SWiG}, \textcolor{blue}{VG}, \textcolor{blue}{COCO} & 52.5 & 52.7 & 52.9 & 50.8 & 53.4 & \underline{47.6} &61.3 &60.9  &54.0\\
VGDiffZero~\cite{liu2023vgdiffzero}  &VAE  &LAION-5B  & 28.0 & 30.3 & 29.1 & 28.4 & 30.8 & 29.8 &33.5  &33.2 &33.9\\
Pseudo-Q$^\dagger$~\cite{jiang2022pseudo}  &R50 &\textcolor{blue}{VG}, \textcolor{blue}{RefC-pseudo}  & 56.0 & 58.3 & 54.1 & 38.9 & 45.1 & 32.1 &49.8 &47.4  &47.7\\
\hline
GLIP~\cite{li2021grounded} &Swin-T &\textcolor{blue}{O365}, \textcolor{blue}{GoldG}, Cap4M & 50.4 & 54.3 & 43.8 & 49.6 & 52.8 &44.6 &\underline{66.1} &\underline{66.9} &53.6 \\
Grounding-Dino~\cite{liu2023grounding} &Swin-T &\textcolor{blue}{O365},\textcolor{blue}{GoldG} & 50.4 & 57.2 & 43.2 & 51.4 & 57.6 &45.8 &\textbf{67.5} &\textbf{67.1} &55.0 \\
MM-G~\cite{zhao2024open} &Swin-T &\textcolor{blue}{O365}, \textcolor{blue}{GoldG}, \textcolor{blue}{GRIT}, \textcolor{blue}{V3Det} & 53.1 & 59.1 & 46.8 & 52.7 & 58.7 &\textbf{48.4} &62.9 &62.9 &55.6 \\
Kosmos-2~\cite{kosmos2}  &ViT-L & KOSMOS-1, \textcolor{blue}{GRIT} &52.3 &57.4 &47.3 &45.5 &50.7 &42.2 &60.6 &61.7 &52.2 \\
CoVLM~\cite{li2024covlm} &ViT-L &Pythia, \textcolor{blue}{CoVLM-97M}  &48.2 &53.2	&43.2	&47.6	&50.9	&44.2	&60.9	&61.9	&51.3 \\
\bf Ours &R50 &\textcolor{blue}{COCO}, \textcolor{blue}{VLM-VG} & \underline{60.8} & \underline{67.1} & \underline{54.9} & 52.5 & 59.6 &43.3 &61.4 &61.2 &\underline{57.6} \\ 
\bf Ours &R101 &\textcolor{blue}{COCO}, \textcolor{blue}{VLM-VG} & \textbf{63.4} & \textbf{68.5} & \textbf{57.6} & \underline{53.9} & \underline{60.9} &44.9 &63.3 &63.2 &\textbf{59.5} \\ 
\bottomrule
\end{tabular}
}
\end{center}
\vspace{-5mm}
\caption{Comparison with state-of-the-art methods on zero-shot referring expression comprehension (REC) tasks on RefCOCO/+/g dataset. Methods shown in top columns take pre-extracted proposals from FRCNN-101 as input, bottom methods take in images. The best two results are \textbf{bold-faced} and \underline{underlined}, respectively. $^*$ denotes methods using model ensembling, Pseudo-Q$^\dagger$ is directly trained on RefCOCO/+/g images, \textcolor{blue}{blue} indicates grounding/localization datasets.}
\label{tab:sota_rec}
\end{table*}

\subsection{Experiment Setting}
To quantitatively evaluate the effectiveness of our human-free grounding annotations, we train the grounding model with constructed dataset for REC and RES task respectively. For REC task, we train on the full-set of VLM-VG combing both COCO and Objects365 splits, for RES task, we only train on VLM-VG-COCO since Objects365 doesn't contain mask annotations. After training, we perform zero-shot REC and RES evaluation on the three popular visual grounding benchmarks: RefCOCO~\cite{refcoco}, RefCOCO+~\cite{refcoco} and RefCOCOg~\cite{refcocog}. For REC task, we report the top-1 accuracy where the predicted box is correct if its IOU with the ground-truth bounding box is greater than 0.5. For the RES task we report the overall IOU (oIOU) and mean IOU (mIOU) following the norms of the community.

\subsection{Referring Expression Comprehension}
\label{sec:rec_exp}
We evaluate and report zero-shot referring expression comprehension performance on RefCOCO/+/g datasets and compare with state-of-the-art zero-shot methods in Table~\ref{tab:sota_rec}. The simple grounding model based on ResNet-101 trained on our proposed VLM-VG dataset achieves best average performance among the three datasets and 8 splits comparing with current SoTA models including two LLM-based models~\cite{kosmos2, li2024covlm}, methods trained on grounding datasets with manual text labeling ~\cite{yao2021cpt, han2023zero, li2021grounded, liu2023grounding, zhao2024open}, and weakly-supervised method~\cite{jiang2022pseudo} utilizing images from RefCOCO/+/g. In particular, on RefCOCO splits which require spatial relationship modeling, our model largely outperforms the SOTA results by up to 7.4 points. On the challenging RefCOCO+ splits, we achieve the second best performance while the SoTA model Red Circle~\cite{shtedritski2023does} uses model ensembling during inference. It's also worth noting that after switching to the smaller ResNet-50 backbone, our method still maintains a highly-competitive performance that is better than all other baseline models including methods take in object proposals from pre-trained Faster R-CNN based on ResNet-101. The strong performance achieved by the light-weight grounding models on the REC task demonstrates the quality of VLM-VG grounding annotations and the effectiveness to scale grounding pre-training with our proposed automatic annotation pipeline.

\subsection{Referring expression Segmentation}
We also conduct zero-shot evaluation on the referring expression segmentation (RES) task and compare with state-of-the-arts models by both mIOU and oIOU metrics on RefCOCO/+/g benchmarks in Table~\ref{tab:sota_res}. We initialize the RES model with the pre-trained REC model from Sec.~\ref{sec:rec_exp} and tune the model on VLM-VG-COCO. We almost achieve best performance on all three datasets and largely outperform the existing methods, which a 5.1\% and 4.1\% of average improvement on oIOU and mIOU respectively with compact model designs, comparing with the state-of-the-arts models~\cite{suo-etal-2023-text} utilizing large-scale segmentation pre-trained SAM~\cite{kirillov2023segment}. Particularly, we observe the similarly strong improvement on RefCOCO as REC results shows, achieving up to 13\% improvement comparing with previous SoTA, further demonstrating the effectiveness of our proposed referring expression annotation pipelines. Similar to REC results, we observe the compact ResNet-50 based model also significantly outperforms existing methods.

\begin{table*}[t]\footnotesize
\begin{center}
\resizebox{2\columnwidth}{!}{
\begin{tabular}{c|c|c|c|ccc|ccc|cc|c}
    \toprule
    \multirow{2}{*}{Metric} &\multirow{2}{*}{Method} &\multirow{2}{*}{Vison Backbone} & \multirow{2}{*}{Pre-training Dataset} &  \multicolumn{3}{c|}{RefCOCO} & \multicolumn{3}{c|}{RefCOCO+} & \multicolumn{2}{c|}{RefCOCOg} & \multicolumn{1}{c}{Avg}\\
    & & & & val & testA & testB & val & testA & testB & val & test &\\

    \midrule
\multirow{10}{*}{oIOU} &Grad-CAM~\cite{yu2023zero} &R50 &\textcolor{blue}{COCO}$^\dagger$, CLIP & 14.0 & 15.1 & 13.5 &14.5 & 15.0 & 14.0 & 12.5 &12.8 &13.9\\
&Score map~\cite{yu2023zero} &R50 &\textcolor{blue}{COCO}$^\dagger$, CLIP & 19.9 & 19.3 & 20.2 &20.4 & 19.7 & 20.8 & 18.9 &19.2 &19.8\\
&Region Token~\cite{yu2023zero} &ViT-B &\textcolor{blue}{COCO}$^\dagger$, CLIP & 21.7 & 20.3 & 22.6 & 22.6 & 20.9 & 23.5 &25.5 &25.4  &22.8\\
&Cropping~\cite{yu2023zero} &R50 &\textcolor{blue}{COCO}$^\dagger$, CLIP & 22.4 & 20.5 & 22.7 & 24.0 & 22.0 & 23.5 &28.2 &27.6  &23.9\\
&Global-Local~\cite{yu2023zero} &R50 &\textcolor{blue}{COCO}$^\dagger$, CLIP & 24.6 & 23.4 & 24.4 & 25.9 & 24.6 & 25.6 &30.1 &29.8  &26.1\\
&SAM-CLIP~\cite{ni2023ref} &ViT-B &\textcolor{blue}{SAM}, CLIP & 25.2 & 25.9 & 24.8 & 25.6 & 27.8 & 26.1 &33.8 &34.8  &28.0\\
&Ref-Diff~\cite{ni2023ref} &VAE &LAION-5B & 35.2 &37.4 &34.50 &35.6 & 38.7 & \bf31.4 &38.6 &37.5  &\underline{36.1}\\
&Tas~\cite{suo-etal-2023-text} & ViT-B&\textcolor{blue}{SAM}, BLIP$^\ddagger$, LAION, CLIP & 29.5 & 30.3 & 28.2 & 33.2 &38.8 & 28.0 &35.8 &36.2  &32.5\\
& \bf Ours &R50 &\textcolor{blue}{COCO}, \textcolor{blue}{VLM-VG} & \underline{43.2} & \underline{46.4} & \underline{39.6} & \underline{35.7} & \underline{39.2} &29.4 &\underline{42.1} &\underline{42.9} &\underline{39.8} \\
& \bf Ours &R101 &\textcolor{blue}{COCO}, \textcolor{blue}{VLM-VG} & \textbf{45.4} & \textbf{48.0} & \bf{41.4} & \textbf{37.0} & \textbf{40.7} &\underline{30.5} &\textbf{42.8} &\bf44.1 &\textbf{41.2} \\
\hline

\multirow{11}{*}{mIOU} &Grad-CAM~\cite{yu2023zero} &R50 &\textcolor{blue}{COCO}$^\dagger$, CLIP & 14.0 & 15.1 & 13.5 &14.5 & 15.0 & 14.0 & 12.5 &12.8 &13.9\\
&Score map~\cite{yu2023zero} &R50 &\textcolor{blue}{COCO}$^\dagger$, CLIP & 14.2 & 15.9 & 13.2 &14.8 & 15.9 & 13.8 & 12.5 &13.2 &14.2\\
&Region Token~\cite{yu2023zero} &ViT-B &\textcolor{blue}{COCO}$^\dagger$, CLIP & 23.4 & 22.1 & 24.6 & 24.5 & 22.6 & 25.4 &27.6 &27.3  &24.7\\
&Cropping~\cite{yu2023zero} &R50 &\textcolor{blue}{COCO}$^\dagger$, CLIP & 24.3 & 22.4 & 24.7 & 26.3 & 23.9 & 25.7 &31.3 &30.9  &26.2\\
&Global-Local~\cite{yu2023zero} &R50 &\textcolor{blue}{COCO}$^\dagger$, CLIP & 26.7 & 25.0 & 26.5 & 28.2 & 26.5 & 27.9 &33.0 &33.1  &28.4\\
&CaR~\cite{sun2023clip}&ViT-B, ViT-L &CLIP & 33.6 & 35.4 & 30.5 & 34.2 & 36.0 & 31.0 &36.7 &36.6  &34.3\\
&SAM-CLIP~\cite{ni2023ref} &ViT-B &\textcolor{blue}{SAM}, CLIP & 26.3 & 25.8 & 26.4 & 25.7 & 28.0 & 26.8 &38.8 &38.9  &29.6\\
&Ref-Diff~\cite{ni2023ref} &VAE &LAION-5B & 37.2 & 38.4 & \underline{37.2} & 37.3 & 40.5 & 33.0 &44.0 &44.5  &39.0\\
&Tas~\cite{suo-etal-2023-text} &ViT-B&\textcolor{blue}{SAM}, BLIP$^\ddagger$, LAION, CLIP & 39.8 &41.1 & 36.2 & \bf43.6 & \bf49.1 & \bf36.5 &46.6 &46.8  &42.5\\
& \bf Ours &R50 &\textcolor{blue}{COCO}, \textcolor{blue}{VLM-VG} & \underline{47.7} & \underline{51.8} & \underline{44.7} & 41.2 & 45.9 &34.7 &\underline{46.6} &\underline{47.1} &\underline{45.0} \\
& \bf Ours &R101 &\textcolor{blue}{COCO}, \textcolor{blue}{VLM-VG} & \bf{49.9} & \bf{53.1} & \bf{46.7} & \underline{42.7} & \underline{47.3} &\underline{36.2} &\bf{48.0} &\bf48.5 &\bf{46.6} \\

\bottomrule
\end{tabular}
}
\end{center}
\vspace{-5mm}
\caption{Comparison with state-of-the-art methods on zero-shot referring expression segmentation (RES) tasks on RefCOCO/+/g dataset. The best two results are \textbf{bold-faced} and \underline{underlined}, respectively. COCO$^\dagger$ means method only uses images from COCO without annotations,  BLIP$^\ddagger$~\cite{li2023blip} contains grounding/localization dataset including COCO and VG, \textcolor{blue}{Blue} indicates grounding/localization datasets.}
\label{tab:sota_res}
\end{table*}

\subsection{Scaling beyond Detection Datasets}
To evaluate the robustness and adaptability of our approach in leveraging generative Vision and Language Models (VLMs) for scalable visual grounding annotation, we further explore the use of images from the WebLI dataset~\cite{chen2023pali} filtered to include images with high-quality text descriptions according to~\cite{align2021}. In WebLI datast, each image is annotated with noisy bounding boxes generated by OWL-ViT~\cite{owlvit2022}. We selected bounding boxes with an OWL-ViT confidence score exceeding 0.8 to ensure a reasonable quality of visual data. From this subset, a total of 100,000 bounding boxes were randomly sampled and subsequently annotated by the PaLI-3~\cite{chen2023pali3} model, which generated captions for these visual regions (type 1 annotations). Table~\ref{tab:ablation_webli} shows the zero-shot referring expression comprehension (REC) accuracy of our model on validation sets of RefCOCO splits. The training settings follow the ablation setups of the main paper (512 batch size for 50K iterations).
The results show significant performance gains with respect to the increase of the dataset size. This demonstrates the promising potential of using web-collected billion-scale data to further scale up visual grounding datasets and models.

\subsection{Ablations and Analysis}
\noindent \textbf{Dataset Design.}
In order to achieve a fine-grained understanding of the contribution of each dataset component, we conduct ablation study to evaluate the generated dataset on REC task step by step. We discuss five dataset variants where each variant is a subset of the next variant from top to down as demonstrated in Table~\ref{tab:ablation_data_design}. 

Initially, we construct dataset variant (a) exclusively using COCO, solely utilizing the Type 1 regional captions, as elaborated in Sec.\ref{sec:dataset_construct}. The grounding model pretrained on variant (a) achieves an average performance of 46.2\%, which stands in comparison with\cite{jiang2022pseudo}, a semi-supervised model trained on RefCOCO/+/g images and pseudo queries generated via rule-based methods. For variant (b), we augment the data source by incorporating images from Objects365 v1 dataset in addition to COCO compared to variant (a). By combining COCO and Objects365, we observe notable improvement across all three datasets, achieving an average performance of 51.7\% with an improvement of 5.5\%. This shows the substantial benefits of scaling grounding pretraining even with out-of-domain image sources such as Objects365 to benefits for model's generalizability. Variant (c) builds upon variant (b) by incorporating horizontal relation expressions (Relation v1) from Type 2, resulting in a significant performance improvement of up to 13 points on RefCOCO splits, which necessitate spatial relationship modeling. Interestingly, we also observe a consistent performance enhancement on RefCOCO+/g splits that don't contain positional references. Build on variant (c), variant (d) adds spatial modeling across vertical and depth dimensions (Relation v2) and further enhance the average performance to 58.7\% and outperforms all the existing SoTA methods. Finally, following Type 3 in Sec.\ref{sec:dataset_construct}, we further investigate whether explicit attribute modeling via prompting the generative VLMs can fix the potential missing details from the ``general region captions“ (type 1). As variant (e) shows, adding attribute-rich descriptions that incorporate supplementary knowledge from human and LLMs only yields limited improvement compared to (d). This suggests that generative VLMs, such as PaLI-3, possess the capability to generate high-quality and detailed descriptions for object-centric image regions that cover typical attributes and benefit scalable and generalizable visual grounding in the wild.

\begin{table*}[t]\footnotesize
\begin{center}
\resizebox{1.9\columnwidth}{!}{
\begin{tabular}{c|c|c|ccc|ccc|cc|c}
    \toprule
    \multirow{2}{*}{Variant} &\multirow{2}{*}{Source Dataset} &\multirow{2}{*}{referring expression} &  \multicolumn{3}{c|}{RefCOCO} & \multicolumn{3}{c|}{RefCOCO+} & \multicolumn{2}{c|}{RefCOCOg} & \multicolumn{1}{c}{Avg}\\
    & & & val & testA & testB & val & testA & testB & val & test &\\

    \midrule
(a)&COCO & Regional Cap. &  43.1 &48.4 &37.1 & 46.0 & 50.4 & 39.8 & 52.3 & 52.3 & 46.2 \\
(b)&COCO, O365 & Regional Cap. &  47.4 &50.9 &42.0 &51.4 &54.5 &45.9 &60.7 & 60.9 & 51.7 \textcolor{teal}{(+5.5)}  \\
(c)&COCO, O365 & Regional Cap. + Rel. v1 &  59.0 &63.1 &52.5 &51.9 &56.7 &44.1 &62.6 &62.8 &56.6 \textcolor{teal}{(+4.9)} \\
(d)&COCO, O365 & Regional Cap. + Rel. v2 &  63.0 &66.3 &\bf58.9 &52.4 &56.7 &\bf46.0 &\bf63.6 &62.8 &58.7 \textcolor{teal}{(+2.1)}\\
(e)&COCO, O365 & Regional Cap. + Rel. v2 + Attr. & \bf63.4 & \bf68.5 & 57.6 & \bf53.9 & \bf60.9 &44.9 &63.3 &\bf63.2 &\textbf{59.5} \textcolor{teal}{(+0.8)} \\

\bottomrule
\end{tabular}
}
\end{center}
\vspace{-6mm}
\caption{\textbf{Ablation on dataset designs}. We compare the influence of the dataset designs including source dataset and referring expression types. In reffering expression, Regional Cap. denotes the Type 1 annotations in Sec.\ref{sec:dataset_construct}, Rel. v1 denotes the horizontal relations in Type 2 and Rel. v2 represents all the three spatial relations. Attr. means the Type 3 attribute-rich descriptions.}
\label{tab:ablation_data_design}
\end{table*}

\begin{table*}[ht]
    \centering
    \begin{tabular}{cc}
        \begin{minipage}{0.5\textwidth}
        \scalebox{0.8}{
            \begin{tabular}{c|c|c|c|c}
                \toprule
                Data size&  RefCOCO & RefCOCO+ & RefCOCOg & Avg\\
                    \midrule
                10k & 19.0 & 19.8 & 21.0 & 20.4 \\
                30k & 23.6 & 24.1 & 26.1 & 25.1 \\
                100k & 26.0 & 27.9 & 30.8 & 29.4 \\
                \bottomrule
                \end{tabular}
            }
        \centering
        \vspace{-3mm}
        \caption{\textbf{REC with WebLI dataset} shows promising potential of further scaling visual grounding with web-collected large-scale data.}
        \label{tab:ablation_webli}
        \end{minipage} &
        \begin{minipage}{0.5\textwidth}
            \scalebox{0.7}{
            \begin{tabular}{c|c|c|c}
                \toprule
                Data Source&  RefCOCO & RefCOCO+ & RefCOCOg\\
            
                \midrule
            Category name & 24.1 & 24.2 & 26.4 \\
            Prompted Caption & 32.2 &34.8 &41.7\\
            No-prompt Caption & 27.3 & 29.6 & 40.1 \\
            Attributes & 34.0 &36.2 &39.0\\
            \bottomrule
            \end{tabular}
        }
        \centering
        \vspace{-3mm}
        \caption{\textbf{Ablation on text annotation source.} Annotation using attributes and prompted captions show competitive performance.}
        \label{tab:ablation_data_source}
        \end{minipage} \\
        \begin{minipage}{0.5\textwidth}
            \vspace{1mm}
            \scalebox{0.7}{
            \begin{tabular}{c|c|ccc|ccc|cc|c}
                \toprule
                \multirow{2}{*}{Metric}&\multirow{2}{*}{REC Pt.} & \multicolumn{3}{c|}{RefCOCO} & \multicolumn{3}{c|}{RefCOCO+} & \multicolumn{2}{c|}{RefCOCOg} & \multicolumn{1}{c}{Avg}\\
                & & val & testA & testB & val & testA & testB & val & test &\\
            
                \midrule
                \multirow{2}{*}{oIOU}& & 43.1 & 43.0 &\bf42.8 & 28.9 & 30.3 &26.4 &38.2 &39.1 &36.5 \\ 
                &\ding{52} &\textbf{45.4} & \textbf{48.0} &41.4 & \textbf{37.0} & \textbf{40.7} &\bf30.5 &\textbf{42.8} &\bf44.1 &\textbf{41.2} \\\midrule
                
                \multirow{2}{*}{mIOU}& & 48.5 & 48.9 & \textbf{49.2} & 35.0 & 37.1 &33.1 &43.5 &44.6 &42.5 \\
                &\ding{52} &\bf49.9 & \bf{53.1} & 46.7 & \bf{42.7} & \bf{47.3} &\bf{36.2} &\bf{48.0} &\bf48.5 &\textbf{46.6}\\
            \bottomrule
            \end{tabular}
            }
            \centering
            \vspace{-3mm}
            \caption{\textbf{Ablation on REC pre-training for RES.} VLM-VG data can benefit the pixel-level task by grounding knowledge transfer.}
            \label{tab:rec_for_res}
        \end{minipage} & 
        \begin{minipage}{0.5\textwidth}
        \vspace{1mm}
        \scalebox{0.7}{
        \begin{tabular}{c|c|c|c|c}
            \toprule
            Mixture Ratio&  RefCOCO & RefCOCO+ & RefCOCOg & Avg\\
            \midrule
            1:1:1:1 & 50.8 &47.7 &55.3 &51.2\\
            1:2:1:2 & 48.6 &46.8 &52.6 &49.4\\
            2:2:1:1 & 52.3 &46.2 &53.5 &50.7\\
            1:1:2:2 & 49.7 &48.4 &55.0 &51.0\\
            2:1:2:1 & 52.2 &47.1 &54.4 &51.3\\
            1:2:1:2 & 48.6 &46.8 &52.6 &49.4\\
            \bottomrule
        \end{tabular}
        }
        \centering
        \vspace{-3mm}
        \caption{\textbf{Ablation on data mixture ratio.} Simple 1:1:1:1 ratio leads to very strong performance.}
        \label{tab:ablation_mix}
        \end{minipage}
    \end{tabular}
\end{table*}

\begin{figure*}[tb]
\centering
\includegraphics[width=0.9\linewidth]{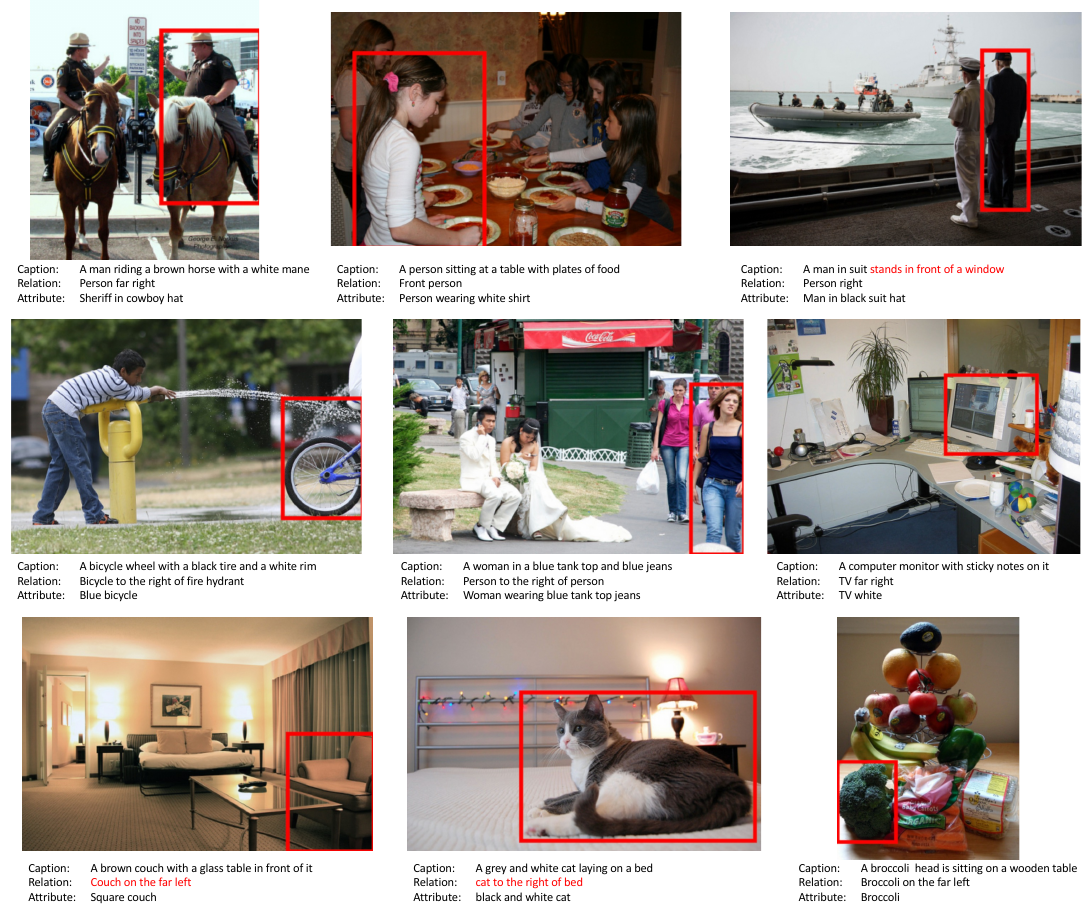}
\caption{\textbf{Visualization of VLM-VG dataset.} By human examination, the incorrect or inaccurate annotations are colored \textcolor{red}{red}.}
\label{fig:vis_data_suppl}
\end{figure*}

\begin{figure*}[tb]
\centering
\includegraphics[width=0.9\linewidth]{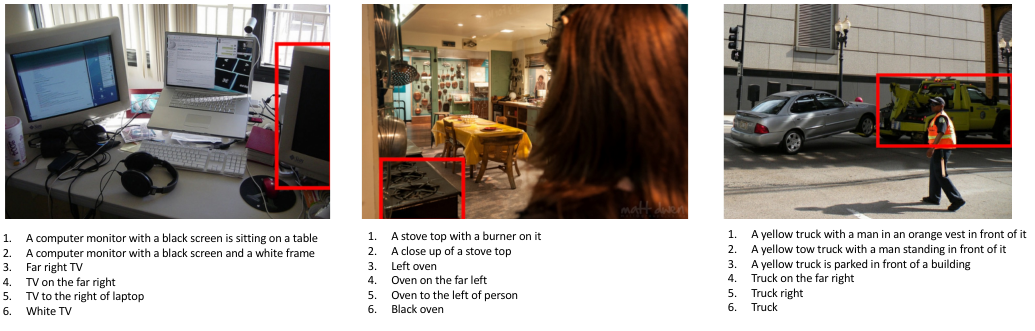}
 \captionsetup{width=0.95\textwidth} 
\caption{\textbf{Diversity of generated annotations.} Our VLM-VG dataset provides referring expressions annotations from multiple perspectives aligning with human linguistic manners.}
\label{fig:vis_data_diversity}
\end{figure*}

\begin{figure*}[tb]
\centering
 \includegraphics[width=0.9\linewidth]{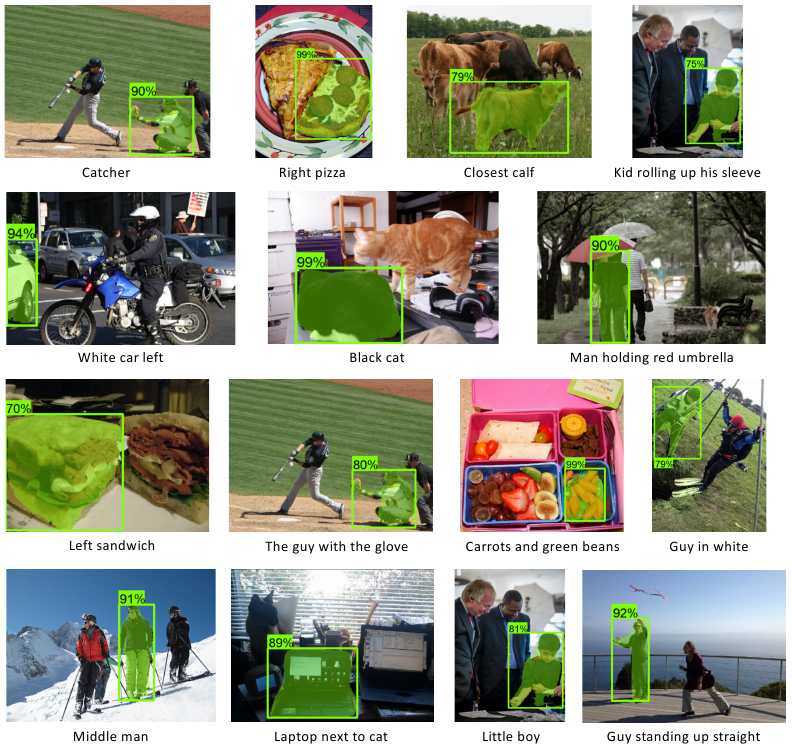}
 \captionsetup{width=0.9\textwidth} 
\caption{\textbf{Visualization of the zero-sot REC and RES predictions on RefCOCO and RefCOCO+.} RefCOCO dataset requires spatial relationship understanding.}
\label{fig:vis_model_pos1}
\vspace{-2mm}
\end{figure*}

\noindent \textbf{Transfer from REC to RES.} Compared with detection dataset, segmentation dataset usually are harder to scale due to the cost of mask annotations. We thus investigate if pre-training on the REC task can further benefit to the RES task. We validate the influence of initializing the RES model with pre-trained REC checkpoints. Results in Table~\ref{tab:rec_for_res} shows that after initializing the model from pre-trained REC checkpoint, the overall RES performance achieves solid improvement especially for the RefCOCO+ splits, this shows VLM-VG data can significantly benefit the pixel-level zero-shot RES task by grounding knowledge transfer. 

\noindent \textbf{Text Annotation Source.}
We conduct ablations on the individual text annotations sources to investigate the influence of referring expression construction methods on COCO dataset. We construct text annotations using 4 different methods: 1) category name directly using class label provided by COCO dataset. 2) prompted caption generated by Type 2 in Sec.~\ref{sec:dataset_construct}. 3) No-prompt caption generated using same method as 2) without providing any language prompts to PaLI-3. 4) Attribute-rich descriptions that constructed by Type 3 in Sec.~\ref{sec:dataset_construct}. We experiment on a small training scale setting (512 batch size for 50K iterations) for efficiency. Results on REC validation set are shown in Table~\ref{tab:ablation_data_source}. We can see directly using category label as the referring expression produce a relative low performance due to the absence of detailed descriptions. Using the regional captions generated by prompting the VLM, the grounding model achieves a huge improvement compared with the category label, demonstrating the abundant additional knowledge gained from the large generative VLMs. Besides, we also compare the influence of prompts during the regional caption generation in line 2 and line 3. We show that designing a proper instruction as the prompt do benefit the quality of generated captions and further enhance the grounding model's performance. Finally, by comparing the second row and the last row, we show that the generic regional captions can achieve similar performance with attribute-rich descriptions involving cross-domain knowledge, further showing the generative VLM's capability to provide informative and comprehensive descriptions of objects by ``generic” instruction prompts.

\noindent \textbf{Data Mixture Ratio.}
We conduct ablations on the data mixture ratios among the constituent datasets of VLM-VG in Table~\ref{tab:ablation_mix}. The mixture ratio is given by (COCO-common, COCO-attribute, Ojbects365-common, Objects365-attributes) where the common datasets denote the combination of datasets from Type 1 and Type 2 in Sec.~\ref{sec:dataset_construct} while attributes dataset represent the attribute-rich descriptions from step 3. We conduct experiment on the same small setting on REC validation set. Results show that a simple 1:1:1:1 ratio leads to very strong performance compared to other ratios, so we adopt it directly.

\noindent \textbf{Inference Time.} We measure the inference time of our model with R50 backbone on 384$\times$384 image to be $60$ms on a single 1080-Ti GPU, which is in the same ballpark of existing REC specialist models~\cite{yang2019fast,deng2021transvg,Luo_2020_CVPR}.

\subsection{Visualization and Analysis}
\noindent \textbf{Dataset.} 
We visualize some examples from the generated VLM-VG dataset with the three types of annotations in Figure~\ref{fig:vis_data_suppl}. By human inspection, the incorrect or inaccurate annotations are labeled as \textcolor{red}{red}. We can see the regional captions generated by the VLM could generally provide detailed and accurate descriptions of the major object, \eg in the third example in the second row, the VLM not only successfully recognizes the object as a computer monitor, but also captures the detail of the sticky notes on the monitor. However, the caption solely relied on cropped regions sometimes might miss or mistakenly describe the global scenes. \eg in the third example in the first row, the caption successfully describes the major object as ``a man in suit” but mistakenly recognizes the action as ``standing in front of a window” since the cropped regions didn't contain the global information such as the ocean and ship. Besides, we also observe that although most of the relation annotations could provide correct spatial information to refer to the object, the simple rule-based method sometimes may still fail to generate the most appropriate spatial descriptions due to the complexity of the scene. 

Besides the high quality of the automatically generated referring annotations, another major advantage of VLM-VG dataset is the diversity of the text annotations. By combing various types of annotations, VLM-VG can annotate one single object from multiple perspectives and at different levels of granularity, which matches human cognition and linguistic manners. Figure~\ref{fig:vis_data_diversity} shows three examples that contain multiple referring annotations varying from the level of detail to the angle of descriptions. Trained on the VLM-VG dataset with diverse annotations, the grounding model can achieve stronger robustness and generalizability.

\begin{figure*}[tb]
\centering
\includegraphics[width=0.8\linewidth]{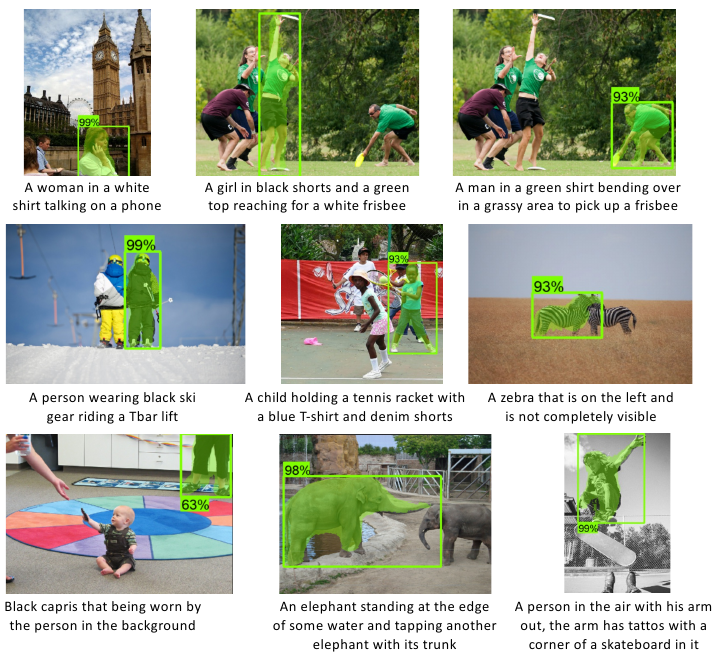}
\vspace{-3mm}
\captionsetup{width=0.9\textwidth} 
\caption{\textbf{Visualization of the zero-sot REC and RES predictions on RefCOCOg.} RefCOCOg requires models to understand longer and more complex referring expressions.}
\label{fig:vis_model_pos2}
\end{figure*}

\begin{figure*}[tb]
\centering
\includegraphics[width=0.8\linewidth]{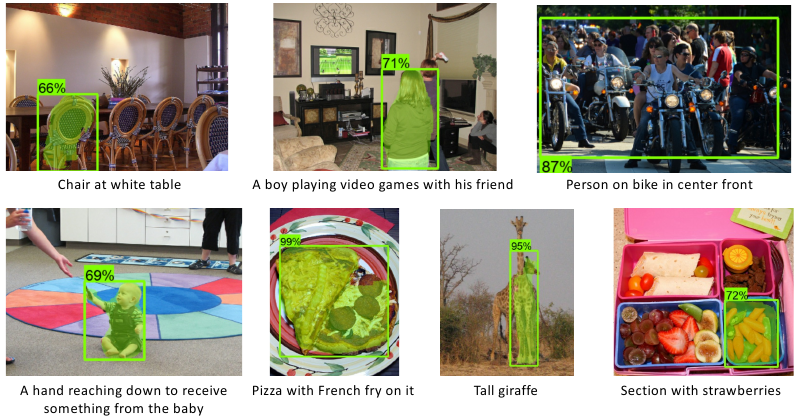}
\vspace{-3mm}
\captionsetup{width=0.8\textwidth} 
\caption{\textbf{Failure cases of the model prediction.}}
\label{fig:vis_model_neg}
\end{figure*}

\noindent \textbf{Model prediction.} 
We illustrate the model's zero-shot REC and RES predictions in Figure~\ref{fig:vis_model_pos1},~\ref{fig:vis_model_pos2}, and~\ref{fig:vis_model_neg}. In detail, Figure~\ref{fig:vis_model_pos1} shows some examples on RefCOCO and RefCOCO+ datasets which use relatively short simple phrases as referring expressions. Trained on VLM-VG dataset, our model can successfully detect objects, understand spatial relations, and distinguish objects by their attributes accurately without seeing human-annotated grounding data. Figure~\ref{fig:vis_model_pos2} shows results on the RefCOCOg dataset which require models to understand longer and more complex sentences as referring expressions. The model also demonstrate a solid capability to associate objects with complex descriptions. For example, the second and third example in Figure~\ref{fig:vis_model_pos2} referring to two people in one image. The model successfully distinguished and located the two people with similar dressing yet different actions, indicating the model's fine-grained reasoning capability.

In order to better understand the shortcomings of the model, we collect several representative failed examples as illustrated in Figure~\ref{fig:vis_model_neg}. One of the major failure modes is that the model fails to capture the visual details mentioned in the referring expression, \eg the ``white table” in the corner in the first image and ``strawberries” in the last images. Moreover, we also observe that complex scenes, such as the one depicted in the third image, pose challenges for the model to locate the correct object by spatial relationships. Furthermore, the third example in the second row reveals a potential limitation of the VLM-VG dataset: it may not cover all the intricate relationships present in real-world scenarios which are hard to be captured by the simple rule-based relation modeling method.

\subsection{Additional Implementation Details}
\label{supp:details}
\noindent \textbf{Relation modeling.} In Section 3.2, after generating the relation tuple \texttt{(noun, rel, noun)} and \texttt{(noun, rel)} for relative and absolute relationship respectively, we use pre-defined templates to formulate the phrases based on the tuple. The templates are listed in Table~\ref{tab:spatial_templates}.

\begin{table}[h]\footnotesize
\begin{center}
\resizebox{\columnwidth}{!}{%
\begin{tabular}{c|c|c}
    \toprule
  Dimension &Tuple & Templates\\

\midrule
\multirow{7}{*}{Horizontal} &\texttt{(A, left, B)} & \textit{A to the left of B}  \\
&\texttt{(A, right, B)} & \textit{A to the right of B} \\
&\texttt{(A, left)} & \textit{A left} / \textit{left A} \\
&\texttt{(A, right)} & \textit{A right} / \textit{right A}\\
&\texttt{(A, left most)} &\textit{A on the far left} / \textit{A far left} / \textit{far left A}\\
&\texttt{(A, right most)} &\textit{A on the far right} / \textit{A far right} / \textit{far right A}\\
&\texttt{(A, middle)} & \textit{A middle} / \textit{middle A} / \textit{center A} / \textit{A center} \\\hline
\multirow{2}{*}{Vertical}&\texttt{(A, top)} & \textit{A top} / \textit{top A} \\
&\texttt{(A, bottom)} & \textit{A bottom} / \textit{bottom A} \\ \hline
\multirow{2}{*}{Depth}&\texttt{(A, behind)} & \textit{A behind} / \textit{behind A} \\
&\texttt{(A, front)} & \textit{A front} / \textit{front A} \\

\bottomrule
\end{tabular}%
}
\end{center}
\vspace{-5mm}
\caption{Templates to formulate spatial relation phrases.}
\label{tab:spatial_templates}
\end{table}

\noindent \textbf{Attributes modeling.} When generating the attribute-rich annotations, we choose 7 types of attributes and query PaLI-3 with the corresponding prompts as detailed in Table~\ref{tab:attr_prompt}. For each attribute, not all the object categories are applicable to the attribute. In details, for the 80 COCO classes, \textit{["cloth", "gender", "identity"]} are applicable to the class \textit{human}, \textit{"action"} is applicable to the class \textit{[person, bird, cat, dog, horse, sheep, cow, elephant, bear, zebra, giraffe]}, \textit{"material"} is applicable to the class \textit{[bench, backpack, umbrella, handbag, tie, suitcase, sports ball, bottle, wine glass, cup, fork, knife, spoon, bowl, chair, couch, bed, dinning table, toilet, sink, clock, boat, vase]}, \textit{"shape"} is applicable to the class \textit{[stop sign, parking meter, bench, handbag, suitcase, kite, bottle, cup, bowl, dining table, couch, bed, toilet, clock, vase]}, and \textit{"color"} is applicable to all the classes. We only query the VLM to model the applicable attributes for each object.

\begin{table}[h]\footnotesize
\begin{center}
\resizebox{0.9\columnwidth}{!}{%
\begin{tabular}{c|c}
    \toprule
Attribute & Prompt\\
\midrule
cloth & \textit{What is the person wearing?}  \\
gender & \textit{What is the person's gender?} \\
identity & \textit{What is the identity of the person?} \\
action & \textit{What is the \{class\} doing?}\\
color & \textit{What is the color of the \{class\}?} \\
material & \textit{What is the material of the \{class\}?}    \\
shape & \textit{What is the shape of the \{class\}?}   \\
\bottomrule
\end{tabular}%
}
\end{center}
\vspace{-5mm}
\caption{Prompts to query PaLI-3 for each attribute. \textit{\{class\}} denotes category name.}
\label{tab:attr_prompt}
\end{table}

\section{Limitations}
When generating the referring expressions, we utilize the rule-based methods utilizing localization heuristics. The manually designed rules as rough approximations for relation on three dimensions empirically show a huge improvement on grounding models' spatial awareness. However, the simple rule-based relation modeling may fail under the complex scenario. For example, when there are adjacent objects with same category, \eg person, the method may produce annotations such as ``person to the left of person” which is not distinctive enough to refer to a specific object. Besides, simply comparing center coordinates and box size to model horizon and depth relation might cause incorrectness due to ignoring the intrinsic size of different objects and struggle with more complex and diverse spatial relations. 

Additionally, we scale grounding datasets based on detection datasets which are generally one to three orders of magnitude larger, without relying on expensive and inflexible manual text annotations. This may be a limitation in the long-run when we want to scale up visual grounding models to objects beyond what are available in current detection datasets. We conducted some initial exploration in Table~\ref{tab:ablation_webli} and observe promising scaling behavior. 

\section{Conclusion}
We present VLM-VG, a large-scale visual grounding dataset built by prompting generative VLM to generate region captions for scaling up visual grounding data. Despite training on image-text data, we observe that existing generative VLMs can produce quality region captions when prompted with object-centric crops from detection datasets and appropriate text instructions. In addition, we propose spatial relation modeling and explicit attribute modeling to capture the linguistic patterns of referring expression. Our VLM-VG dataset include 500K images, 1M objects, and 16M text queries. It is one of the largest grounding datasets to date, and the first with purely machine-generated texts. We demonstrate the advantage of VLM-VG by zero-shot transfer to referring expression comprehension (REC) and segmentation (RES) tasks on RefCOCO benchmarks. Our simple, lightweight model significantly outperforms the state-of-the-art approaches trained on more human-annotated data on both tasks. We hope these findings encourage the community to explore generative VLM for scaling up visual grounding dataset in the real world.

\clearpage
\newpage
{\small
\bibliographystyle{ieee_fullname}
\bibliography{egbib}
}
\end{document}